\documentclass[a4paper,fleqn]{cas-sc}
\usepackage[utf8]{inputenc}
\UseRawInputEncoding

\usepackage[authoryear,longnamesfirst]{natbib}
\usepackage{microtype}
\usepackage{multirow}
\usepackage{colortbl}
\def\tsc#1{\csdef{#1}{\textsc{\lowercase{#1}}\xspace}}
\tsc{WGM}
\tsc{QE}
\tsc{EP}
\tsc{PMS}
\tsc{BEC}
\tsc{DE}

\begin{document}
\let\WriteBookmarks\relax
\def\floatpagepagefraction{1}
\def\textpagefraction{.001}
\shorttitle{}
\shortauthors{Bin Wang et~al.}

\title [mode = title]{TimeCF: A TimeMixer-Based Model with adaptive Convolution and Sharpness-Aware Minimization Frequency Domain Loss for long-term time seris forecasting}

\author[1]{Bin Wang}[orcid=0009-0002-0536-2898]

\ead{wb_csut@csu.edu.cn}

\affiliation[1]{organization={School of Computer Science and Engineering},
                addressline={Central South University}, 
                city={ChangSha},
                postcode={410000}, 
                state={HuNan},
                country={China}}

\author[1]{Heming Yang}[orcid=0009-0009-1683-6068]
\ead{244712142@csu.edu.cn}

\author[1]{Jinfang Sheng}[orcid=0000-0002-6533-7822]
\cormark[1]

\ead{jfsheng@csu.edu.cn}

\cortext[cor1]{Corresponding author}

\begin{abstract}
  Recent studies have shown that by introducing prior knowledge, multi-scale analysis of complex and non-stationary time series in real environments can achieve good results in the field of long-term forecasting. However, affected by channel-independent methods, models based on multi-scale analysis may produce suboptimal prediction results due to the autocorrelation between time series labels, which in turn affects the generalization ability of the model. To address this challenge, we are inspired by the idea of sharpness-aware minimization and the recently proposed FreDF method and design a deep learning model TimeCF for long-term time series forecasting based on the TimeMixer, combined with our designed adaptive convolution information aggregation module and Sharpness-Aware Minimization Frequency Domain Loss (SAMFre). Specifically, TimeCF first decomposes the original time series into sequences of different scales. Next, the same-sized convolution modules are used to adaptively aggregate information of different scales on sequences of different scales. Then, decomposing each sequence into season and trend parts and the two parts are mixed at different scales through bottom-up and top-down methods respectively. Finally, different scales are aggregated through a Feed-Forward Network. What's more, extensive experimental results on different real-world datasets show that our proposed TimeCF has excellent performance in the field of long-term forecasting.
\end{abstract}


\maketitle

\section{Introduction}
With the development of information technology in the past decade, time series forecasting, a field of great significance to human life, has been supported by information technology resources such as computing power and algorithms and has played an indispensable role in key areas related to human living standards, such as financial level prediction(\cite{sonkavde_forecasting_2023}), traffic flow planning(\cite{huo_hierarchical_2023,huang_interpretable_2023}), weather forecast(\cite{bi_accurate_2023}), water treatment(\cite{farhi_prediction_2021,afan_data-driven_2024}), energy and power resource allocation(\cite{alkhayat_review_2021,yin_weighted_2023}). Since the begin of time series forecasting, there are mainly the following model architectures: Models based on CNN(\cite{li_lagcnn_2024}), Models based on RNN(\cite{salinas_deepar_2020}), Models based on Transformer(\cite{liang_crossformer_2024}) and Models based on MLP(\cite{challu_nhits_2023}).

Although researchers have proposed a variety of methods to solve the problem of time series forecasting, the process of capturing and building a model of time series from the past to the future is challenging because the natural sequence of time series has complex and non-stationary properties and the noise from the data acquisition equipment can affect the prediction results. In order to solve this problem, the current mainstream research can be divided into two categories: one is based on the Transformer(\cite{vaswani_attention_2017}), which achieves the fitting of time series through a large number of parameters. However, although the large number of parameters of the Transformer can solve the problem of complex and non-stationary time series to some extent, its easy overfitting and slow training speed have not been reliably solved. Therefore, after fully studying the components of time series, researchers use the prior knowledge of physics and mathematics to decompose time series into simpler components to reduce the difficulty of the prediction process. On this basis, TimeMixer(\cite{wang_timemixer_2024}) further introduced the idea of multi-scale decomposition, and TimeKAN(\cite{huang_timekan_2025}) proposed modeling based on frequencies of different scales based on TimeMixer. In summary, current researchers hope to simplify the original time series to provide additional prior information for the time series model, thereby imporving the prediction accuracy of the time series forecasting model.

It is undeniable that the model based on the idea of channel independence does obtain more accurate results under certain conditions, but the actual time series is a series with high autocorrelation. This autocorrelation is manifested in that it is not only correlated in the order of time, but also has a certain degree of correlation between the labels of the sequence. Therefore, under the premise of channel independence, the autocorrelation between labels has not been fully processed, which may cause a certain degree of distortion in the results of the model. Fortunately, a method called FreDF(\cite{wang_fredf_2024}) has recently been applied to the field of time series forecasting. It uses Fourier or fast Fourier transform to transform the sequence labels from the time domain to the frequency domain without changing the model structure to deal with the autocorrelation in the time series. In addition, we note that an idea called sharpness-aware minimization(SAM)(\cite{foret_sharpness-aware_2021}) can be combined with FreDF to reduce the sharpness of the loss so that the model has better generalization ability. At the same time, inspired by the idea of attention mechanism in Transformer and the idea of receptive field in CNN field, we found that neighboring and global information can also be used in time series to supplement information in the prediction process. Therefore, we propose to use convolution kernels of the same size at different scales to obtain global information at low-frequency scales and neighboring information at high-frequency scales to achieve the aggregation of global and local information with a small number of parameters.

Combining the advantages of the above technologies, we propose a frequency-independent multi-scale hybrid architecture (TimeCF) based on the TimeMixer model to solve the problems of global and local information loss, autocorrelation between sequence labels in time series forecasting and generalization ability. In terms of model structure, TimeCF is based on the TimeMixer model architecture. First, it uses the downsampling method to generate time series at multiple scales. Secondly, through the PDMC (Past Decomposable Mixing with adaptive Conv) module designed by us, we first use the convolution operation of the same convolution kernel on the sequences of different scales to achieve adaptive information aggregation between different scales. Then, according to prior knowledge, the season and trend of the input sequence are decomposed separately. Through our design, PDMC obtains information of different receptive fields according to different input scales and decomposes the sequences of different scales into seasonal and trend parts to achieve more detailed modeling. In the prediction stage, the output prediction layer aggregates the prediction components of different scales to utilize the complementary prediction capabilities between multi-scale sequences to achieve accurate prediction.

In general, our contributions are as follows:
\begin{enumerate}
  \itemsep=0pt
  \item Different from previous methods, we propose to use adaptive convolution modules to achieve information aggregation of receptive fields of different scales based on sampling results of different scales. What's more, we use the transformation from time domain to frequency domain to solve the challenges brought by complex information coupling in time series.
  \item We proposed a relatively lightweight time series prediction model TimeCF and introduced the receptive field idea in the CNN field to maximize the use of information aggregation at different scales to supplement global and local information. And based on the ideas of FreDF and SAM, we achieved the decoupling of the autocorrelation between the labels of the time series and the improvement of the model's generalization ability.
  \item TimeCF shows excellent performance in multiple time series forecasting tasks and datasets, while achieving a relatively balanced state between model parameters and prediction accuracy.
  \end{enumerate}  

\section{Related Work}
\subsection{Mainstream Model Architecture}
The core of the time series forecasting model is to have efficient and stable pattern extraction and modeling capabilities in different time series, so as to model and predict complex time series. Traditional models such as ARIMA(\cite{zhang_time_2003}) and LSTM(\cite{hochreiter_long_1997}) can accurately predict time series with simple cycles and trends, but these models are limited by parameters and model structures so the prediction effect for nonlinear and dynamic time series is often unsatisfactory. In recent years, deep learning methods have begun to make great strides in the direction of time series forecasting. For the Transformer, researchers have proposed many methods to apply it to the field of time series prediction: Autoformer(\cite{wu_autoformer_2021}) proposed an autocorrelation mechanism to reduce the time complexity of the model to $O\left(n\cdot l g\left(n\right)\right)$. SAMformer(\cite{ilbert_samformer_2024}) solved the instability problem during large model training by using Sharpness-Aware Minimization. Informer(\cite{zhou_informer_2021}) used ProbSpare self-attention and Self-attention Distilling to enable it to effectively handle overly long input sequences. iTransformer(\cite{liu_itransformer_2024}) inverted the time series and then used the Encoder for prediction. Mamba(\cite{gu_mamba_2023}) combined the parallelization capability of Transformer and the historical information control capability of RNN, and based on the idea of SSM, it was able to handle the correlation problem between variables at a lower cost. PatchTST(\cite{nie_time_2023}) regarded the time series as multiple independent time periods of channels, and combined it with Transformer for prediction, achieving good results. And some researchers have found that the use of CNN ideas can better construct the relationship between labels and time steps in time series: MICN(\cite{wang_micn_2023}) introduces the idea of image processing and captures information of different receptive fields through convolution kernels of different sizes. TimesNet(\cite{wu_timesnet_2023}) performs Fourier transform on the time series and selects its Top-k cycles, then expands each cycle into a two-dimensional image and uses a 2D-kernel convolution kernel for feature extraction. ModernTCN(\cite{donghao_moderntcn_2024}) proposes to use large convolution kernels on the time dimension of the time series so that the model can capture dependencies across time and variables at the same time. In addition, researchers have also proposed some models that are not limited to Transformer and CNN: GRU(\cite{chung_empirical_2014}) introduces a gating mechanism that allows the model to dynamically adjust the ratio of memory and forgetting according to the current input and previous state, making it more flexible and expressive than traditional RNN. DLinear(\cite{zeng_are_2023}) decomposes time series into seasonal and trend components for separate predictions. FITS(\cite{xu_fits_2024}) roposes the use of basic MLP for prediction in the frequency domain based on Dlinear. And SparseTSF(\cite{lin_sparsetsf_2024}) obtain information about adjacent time steps through convolution, and then predict future results separately through sparse technology.

Considering the advantages and limitations of the above models, people need a time series forecasting model that can extract different features and have accurate prediction results. Therefore, we proposed TimeCF, based on the original idea of scale decomposition, to obtain features of different scales through convolution to achieve multi-scale adaptive information aggregation.
\subsection{Parameters Update}
Nowadays researchers have begun to find that the effects that the models can achieve on the training set are often not achieved on the test set. This is because when the model uses an optimizer to optimize the non-convex loss function on the training set, it may enter a suboptimal or sharp minimum, resulting in insufficient generalization of the model. In response to this situation, researchers have proposed a method called sharpness-aware minimization to update model parameters to improve the generalization ability of deep neural networks: SAM(\cite{foret_sharpness-aware_2021}) proposed the sharpness-aware minimization method, which first finds the point with the maximum loss in the neighborhood of the current parameter and then uses gradient descent to update the parameters based on this maximum point, so that the parameters can be moved to a flat area to reduce the sharpness of the loss function. WSAM(\cite{yue_sharpness-aware_2023}) introduces the concept of weights based on SAM, and adjusts the contribution of different parameters to sharpness according to the importance of the parameters or other indicators, thereby more effectively regularizing. FSAM(\cite{li_friendly_2024}) effectively improves the generalization performance and robustness of the model by improving adversarial perturbations and optimizing the full gradient estimation method.

Considering the generalization degree of the model, we propose a model parameter update module called SAMFre based on the SAM idea to improve the overall generalization ability of the model.

\section{TimeCF}
\subsection{OverView}
The basic definition of time series forecasting is to input the historical data of a multivariate time series $ X_{input} \in R^{T\times N} $ and after the model calculation, output the future multivariate output sequence $X_{output}\in R^{F\times N}$, where T represents the lookback length of the historical data defined by the model, F represents the future time length to be predicted and N represents the number of labels in the time series.

In TimeCF, we use the idea of channel independence to make independent predictions for each label in the time series. Therefore, the original input can be regarded as $ \left\{X_{input^1},X_{input^2},\ldots,X_{input^N}\right\}$ , where $ X_{input^i}\in R^T$ can be regarded as the input instance of TimeCF. The overall structure of TimeCF is shown in Figure 1, which consists of three components: Input Preprocessing layer, PDMC layer, and Output Predicting layer. At the same time, SAMFre as a module to solve the autocorrelation between variables and improve the generalization ability of the model indirectly participates in model training in the stages of calculating model loss and updating model parameters. In summary, the overall process of TimeCF consists of three explicit modules and one implicit module.

\subsection{Input Preprocessing layer}
Since we treat the time series $X_{input^i}\in R^T$ in each label as a separate input instance, for each instance $X_{input^i}$,we first use the pooling layer to generate multi-level sequences of different scales $\left\{X_1,X_2,\ldots,X_k\right\}$, where $X_i\in R^\frac{T}{d^i-1}\left(i\in\left\{1,\ldots k\right\}\right)$. The output $X_i$ is the result of $i-1$ times of downsampling of the original input $ X_{input^i} $. $ X_1 $is equal to the original input sequence $ X_{input^i} $and $d$ represents the length of the moving window in the pooling layer. The specific multi-scale sequence generation formula is as follows:

\begin{eqnarray}
  X_i=Pool\left(Padding\left(X_{\left(i-1\right)}\right)\right)
\end{eqnarray}

After generating the multi-scale sequences, each sequence will have a time-related mask $X_{{\mathrm{mask}}^i}$. Each sequence is first normalized by the RevIn normalization layer, and then the mask and sequence are embedded by the Embedding layer. The specific process is as follows:

\begin{eqnarray}
  X_i=TemporalEmbedding\left(X_{{\mathrm{mask}}^i}\right)+TokenEmbedding\left({RevIN(X}_i)\right)
\end{eqnarray}

In formula (2), the sequence of each scale is $X_i\in R^{\frac{T}{d^i-1}\times D}$, $D$ is the output dimension of embedding. At this point, the preprocessing part of the input data is completed, and this stage is only performed once during the model training process.

\subsection{Past Decomposable Mixing with adaptive Conv layer}
\begin{figure}
	\centering
	\includegraphics[width=1.0\textwidth]{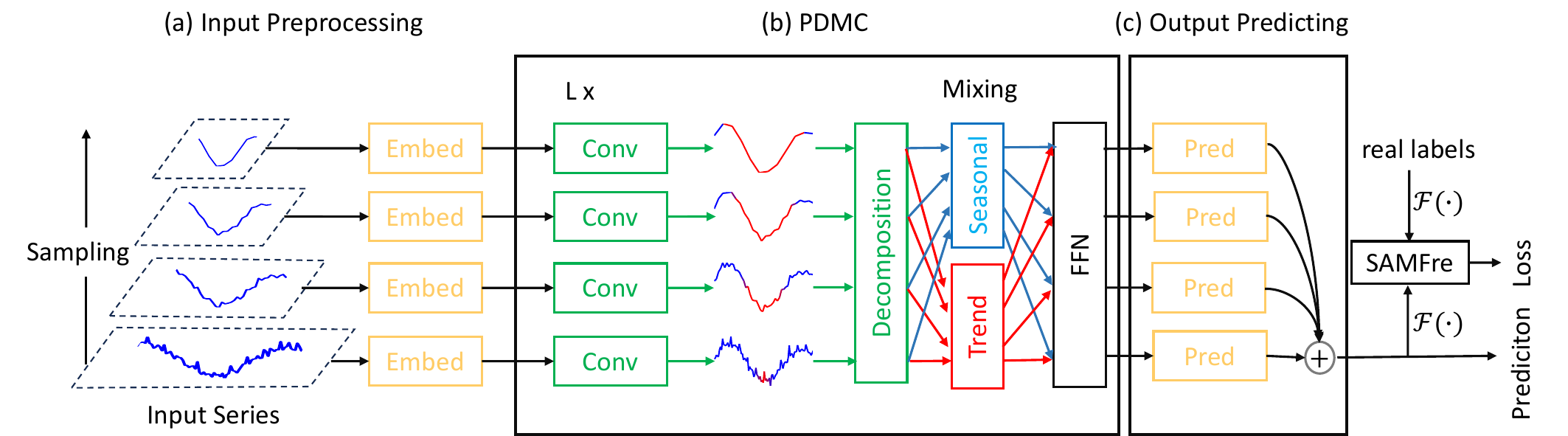}
	\caption{TimeCF Architecture}
	\label{FIG:1}
\end{figure}

Recent studies have found that most time series are the fusion of different components of different periods at most scales. Therefore, we propose the PDMC module, which uses long-term and short-term changes to analyze various periodic and non-periodic properties of the entire time series, while obtaining information of different receptive fields at different scales through convolution. Specifically, in the PDMC, we first add global or local information to the sequence through the idea of convolution and adaptation:

\begin{eqnarray}
  X_i=\alpha\times ConvBlocks_{\left[i\right]}\left(X_i^T\right)^T+X_i
\end{eqnarray}

The $X_i$ is the output of the input preprocessing layer and the formula for $ConvBlocks_{\left[i\right]}\left(X\right)$ is as follow:

\begin{eqnarray}
  ConvBlocks_{\left[i\right]}\left(X\right)=Conv\left(GELU\left(Conv\left(GELU\left(Conv\left(Norm\left(X\right)\right)\right)\right)\right)\right)
\end{eqnarray}

Next, we will explain why using the same size of convolution can obtain information of different receptive fields. First, the lookback window length selected by this model is 96, which is consistent with the mainstream model, and the number of downsampling is set to 3. So the input of PDMC is $\left\{X_1\in R^{96\times D},X_2\in R^{48\times D},X_3\in R^{24\times D},X_4\in R^{12\times D}\right\}$, where 96,48,24,12 are the time windows after downsampling. And in $ConvBlocks_{\left[i\right]}\left(X\right)$,we use three layers of convolution and the convolution kernel size of each layer is 3, and the padding is 1. This means that after three convolutions, each time point in $\left\{X_1,X_2,X_3,X_4\right\}$ contains information from at least $2\ +\ 2\ +\ 2$, or 6 neighboring time points. From the perspective of PDMC stacking, the number of PDMC stackings is $ L(L\geq 2)$. So for $X_4$, the window of length $6\times L $ will eventually cover the entire sequence length, which can be considered as obtaining global information. But for $X_1$,$X_2$ and $X_3$,the window of length $6\times L$ only occupies a part of the sequence length, which can be considered as obtaining local information of different proportions. In summary, by using convolution blocks and PDMC stacking, TimeCF can obtain information of different receptive fields at different scales.

Then, we decompose the sequence of each scale into season and trend parts:

\begin{eqnarray}
  Season_i,Trend_i=Decomp\left(X_i\right)
\end{eqnarray}

$Season_i$,$Trend_i$ refer to the season and trend parts decomposed from the i-th scale respectively. We put all the season and trend components into the lists $Season$ and $Trend$ respectively, and based on the idea of TimeMixer, we perform scale fusion on the season and trend respectively:

\begin{eqnarray}
  Season,Trend=SeasonMix\left(Season\right),TrendMix\left(Trend\right)
\end{eqnarray}

The fusion of the season term is a bottom-up sequence fusion and the trend term is a top-down sequence fusion which make full use of the information inherent in both parts. Finally, PDMC passes the season part, trend part and original sequence through the feed forward network to achieve the fusion between different components:

\begin{eqnarray}
  X_i=X_i+FFN\left(Season_i+Trend_i\right)
\end{eqnarray}

So far, the PDMC block has finally realized the core tasks of feature extraction and multi-scale mixing process through the adaptive information aggregation by convolution, the decomposition and mixing of the season term and trend term.

\subsection{Output Predicting layer}
In the prediction output stage, the output of PDMC we obtain is $\left\{X_1,X_2,\ldots,X_k\right\}$, where $X_i\in R^{\frac{T}{d^i-1}\times D}\left(i\in\left\{1,\ldots k\right\}\right)$. So if we need to make a prediction for $X_i$ in the time dimension, we need to change the dimension of $X_i$ at least twice. Specifically, first align the time dimension of $X_i$ with the predicted future length according to different scales. Then adjust the dimension of the sequence so that the model vector dimension D can be reduced back to the initial value:

\begin{eqnarray}
  X_i=Linear_2\left(Linear_1\left(X_i^T\right)^T\right)
\end{eqnarray}

The input dimension of the $Linear_1$ is $\frac{T}{d^i-1}$, and the output dimension is the prediction length$ F$. As a result, time series of different scales generate predictions of corresponding time lengths. Then, the input dimension of the $Linear_2$ is $D$ and the output dimension 1. This is to make the sequence dimension match the target output dimension, or let $X_i\in R^F$

It is not difficult to see that each scale sequence eventually generates a prediction sequence. Then we sum all the prediction sequences and use the RevIN layer of the preprocessing layer to perform inverse normalization:

\begin{eqnarray}
  X_O=iRevIN\left(\sum X_i\right)
\end{eqnarray}

At this point, the prediction of a single label is completed, and the sequences of different scales are finally fused together through the stack() function to forecast the result.

\subsection{Sharpness-Aware Minimization Frequency Domain Loss}
The loss function of traditional time series forecasting model is usually MSE loss, which has shown its superiority in the training process of a large number of time series forecasting models. However, with the introduction of the idea of channel independence, FreDF's researchers have noticed that the MSE loss hardly takes into account the autocorrelation between different labels of the time series in the model using the channel independence method. Therefore, it is not the best choice to calculate the loss by MSE in the training process of the time series forecasting model using the channel independence method. However, according to the idea of Fourier transform, if different labels are projected into the frequency domain, unrelated feature can be obtained in the frequency domain so that the model based on the idea can obtain better results than the traditional MSE loss when calculating the loss. At the same time, we noticed that the overall generalization performance of the model can be improved by adjusting the sharpness of the loss through the SAM method. Based on these two ideas, the TimeCF we proposed introduces the SAMFre module to decouple the autocorrelation between different labels in the time series and improve the generalization ability. Specifically, SAMFre projects the model's prediction results and the actual label values into the frequency domain through Fourier transform, then calculates the loss using the L1 norm, and finally adds it to the original MSE loss to get the complete loss:

\begin{eqnarray}
  loss=\alpha\times\left|FFT\left(pred\right)-FFT\left(real\right)\right|_1+\left(1-\alpha\right)\times MSE
\end{eqnarray}

After calculating the loss, the model uses basic optimization methods to optimize the model parameters before the number of updates reaches the set threshold. When the number of updates reaches the threshold, the model uses the SAM method to calculate the point with the largest loss in the neighborhood of the current parameter, and then performs gradient backpropagation based on this point to achieve parameter update:

\begin{eqnarray}
  \hat{\epsilon}\left(w\right)\ =\ \ \rho \frac{\nabla_w Loss}{\|\nabla_w Loss\|_2}
\end{eqnarray}
\begin{eqnarray}
  g\ =\ \left.\nabla_wLoss\right|_{w+\hat{\epsilon}\left(w\right)\ }
\end{eqnarray}
\begin{eqnarray}
  w\ =\ w\ -\ \eta\ \cdot\ g
\end{eqnarray}
  
So far, we have optimized the model parameter update part through SAMFre, so that the model can better deal with the autocorrelation problem between labels in different sequences and improve the generalization ability of the model.

\section{Results}
\begin{table*}[!t]
  \centering
  \caption{Performance comparison of different time series forecasting models on benchmark datasets. }
  \label{tab:comparison}
  \resizebox{\textwidth}{!}{%
    \setlength{\tabcolsep}{2pt}
    \begin{tabular}{cccccccccccccccccccc}
    \toprule
    \multicolumn{2}{c}{\multirow{2}{*}{Models}} & \multicolumn{2}{c}{TimeCF} & \multicolumn{2}{c}{TimeKAN} & \multicolumn{2}{c}{TimeMixer} & \multicolumn{2}{c}{iTransformer} & \multicolumn{2}{c}{SparseTSF} & \multicolumn{2}{c}{FreTS} & \multicolumn{2}{c}{PatchTST} & \multicolumn{2}{c}{TimesNet} & \multicolumn{2}{c}{Dlinear} \\
    \multicolumn{2}{c}{} & \multicolumn{2}{c}{Ours} & \multicolumn{2}{c}{2025} & \multicolumn{2}{c}{2024} & \multicolumn{2}{c}{2024} & \multicolumn{2}{c}{2024} & \multicolumn{2}{c}{2024} & \multicolumn{2}{c}{2023} & \multicolumn{2}{c}{2023} & \multicolumn{2}{c}{2023} \\
    \multicolumn{2}{c}{Metric} & MSE & MAE & MSE & MAE & MSE & MAE & MSE & MAE & MSE & MAE & MSE & MAE & MSE & MAE & MSE & MAE & MSE & MAE \\
    \midrule
    \multirow{5}{*}{ETT h1} & 96 & {\color[HTML]{FF0000} \textbf{0.359}} & {\color[HTML]{FF0000} \textbf{0.391}} & \textbf{0.367} & 0.394 & 0.381 & 0.398 & 0.394 & 0.409 & 0.385 & \textbf{0.391} & 0.395 & 0.407 & 0.376 & 0.397 & 0.389 & 0.411 & 0.396 & 0.410 \\
    & 192 & {\color[HTML]{FF0000} \textbf{0.401}} & {\color[HTML]{FF0000} \textbf{0.419}} & \textbf{0.414} & \textbf{0.419} & 0.441 & 0.430 & 0.448 & 0.441 & 0.434 & 0.420 & 0.490 & 0.477 & 0.426 & 0.432 & 0.439 & 0.441 & 0.445 & 0.440 \\
    & 336 & {\color[HTML]{FF0000} \textbf{0.440}} & \textbf{0.436} & \textbf{0.445} & {\color[HTML]{FF0000} \textbf{0.434}} & 0.500 & 0.459 & 0.492 & 0.465 & 0.476 & 0.439 & 0.510 & 0.480 & 0.469 & 0.457 & 0.493 & 0.470 & 0.487 & 0.465 \\
    & 720 & 0.466 & \textbf{0.462} & {\color[HTML]{FF0000} \textbf{0.451}} & 0.463 & 0.552 & 0.507 & 0.521 & 0.504 & \textbf{0.461} & {\color[HTML]{FF0000} \textbf{0.454}} & 0.568 & 0.538 & 0.518 & 0.504 & 0.516 & 0.494 & 0.512 & 0.510 \\
    & avg & {\color[HTML]{FF0000} \textbf{0.417}} & \textbf{0.427} & \textbf{0.419} & 0.428 & 0.468 & 0.449 & 0.464 & 0.455 & 0.439 & {\color[HTML]{FF0000} \textbf{0.426}} & 0.490 & 0.475 & 0.447 & 0.447 & 0.459 & 0.454 & 0.460 & 0.456 \\
    \midrule
    \multirow{5}{*}{ETT h2} & 96 & {\color[HTML]{FF0000} \textbf{0.282}} & {\color[HTML]{FF0000} \textbf{0.333}} & 0.291 & 0.340 & \textbf{0.286} & \textbf{0.339} & 0.300 & 0.349 & 0.302 & 0.346 & 0.332 & 0.387 & 0.308 & 0.359 & 0.337 & 0.370 & 0.341 & 0.395 \\
    & 192 & {\color[HTML]{FF0000} \textbf{0.372}} & {\color[HTML]{FF0000} \textbf{0.389}} & \textbf{0.374} & \textbf{0.391} & 0.391 & 0.404 & 0.381 & 0.399 & 0.384 & 0.395 & 0.451 & 0.457 & 0.380 & 0.406 & 0.404 & 0.414 & 0.481 & 0.479 \\
    & 336 & {\color[HTML]{FF0000} \textbf{0.410}} & {\color[HTML]{FF0000} \textbf{0.422}} & 0.423 & 0.434 & 0.421 & 0.432 & 0.423 & 0.432 & 0.421 & \textbf{0.427} & 0.466 & 0.473 & \textbf{0.412} & 0.429 & 0.455 & 0.452 & 0.592 & 0.542 \\
    & 720 & {\color[HTML]{FF0000} \textbf{0.416}} & {\color[HTML]{FF0000} \textbf{0.436}} & 0.462 & 0.461 & 0.468 & 0.468 & 0.426 & 0.445 & \textbf{0.420} & \textbf{0.437} & 0.485 & 0.471 & 0.435 & 0.456 & 0.434 & 0.448 & 0.840 & 0.661 \\
    & avg & {\color[HTML]{FF0000} \textbf{0.370}} & {\color[HTML]{FF0000} \textbf{0.395}} & 0.387 & 0.406 & 0.391 & 0.411 & 0.383 & 0.406 & \textbf{0.382} & \textbf{0.401} & 0.433 & 0.447 & 0.384 & 0.412 & 0.407 & 0.421 & 0.564 & 0.519 \\
    \midrule
    \multirow{5}{*}{ETT m1} & 96 & {\color[HTML]{FF0000} \textbf{0.307}} & {\color[HTML]{FF0000} \textbf{0.345}} & \textbf{0.321} & \textbf{0.361} & 0.327 & 0.364 & 0.341 & 0.376 & 0.356 & 0.375 & 0.337 & 0.374 & 0.323 & 0.364 & 0.333 & 0.375 & 0.345 & 0.373 \\
    & 192 & {\color[HTML]{FF0000} \textbf{0.353}} & {\color[HTML]{FF0000} \textbf{0.372}} & \textbf{0.356} & \textbf{0.382} & 0.367 & 0.386 & 0.380 & 0.394 & 0.394 & 0.392 & 0.382 & 0.398 & 0.371 & 0.391 & 0.407 & 0.413 & 0.381 & 0.391 \\
    & 336 & {\color[HTML]{FF0000} \textbf{0.377}} & {\color[HTML]{FF0000} \textbf{0.395}} & \textbf{0.381} & \textbf{0.400} & 0.393 & 0.403 & 0.419 & 0.418 & 0.425 & 0.413 & 0.420 & 0.423 & 0.398 & 0.408 & 0.413 & 0.421 & 0.415 & 0.415 \\
    & 720 & {\color[HTML]{FF0000} \textbf{0.441}} & {\color[HTML]{FF0000} \textbf{0.430}} & \textbf{0.451} & \textbf{0.437} & 0.451 & 0.442 & 0.486 & 0.455 & 0.487 & 0.448 & 0.490 & 0.471 & 0.457 & 0.444 & 0.503 & 0.467 & 0.472 & 0.450 \\
    & avg & {\color[HTML]{FF0000} \textbf{0.370}} & {\color[HTML]{FF0000} \textbf{0.386}} & \textbf{0.377} & \textbf{0.395} & 0.384 & 0.399 & 0.406 & 0.411 & 0.415 & 0.407 & 0.407 & 0.416 & 0.387 & 0.402 & 0.414 & 0.419 & 0.403 & 0.407 \\
    \midrule
    \multirow{5}{*}{ETT m2} & 96 & {\color[HTML]{FF0000} \textbf{0.169}} & {\color[HTML]{FF0000} \textbf{0.252}} & 0.175 & 0.257 & \textbf{0.174} & \textbf{0.257} & 0.183 & 0.266 & 0.184 & 0.267 & 0.186 & 0.275 & 0.184 & 0.267 & 0.189 & 0.266 & 0.193 & 0.292 \\
    & 192 & \textbf{0.238} & {\color[HTML]{FF0000} \textbf{0.299}} & 0.239 & 0.299 & {\color[HTML]{FF0000} \textbf{0.236}} & \textbf{0.299} & 0.252 & 0.312 & 0.248 & 0.305 & 0.259 & 0.323 & 0.246 & 0.304 & 0.252 & 0.307 & 0.284 & 0.361 \\
    & 336 & {\color[HTML]{FF0000} \textbf{0.296}} & {\color[HTML]{FF0000} \textbf{0.335}} & \textbf{0.301} & 0.340 & 0.301 & \textbf{0.339} & 0.314 & 0.351 & 0.307 & 0.342 & 0.349 & 0.386 & 0.311 & 0.348 & 0.321 & 0.349 & 0.384 & 0.429 \\
    & 720 & \textbf{0.400} & {\color[HTML]{FF0000} \textbf{0.393}} & {\color[HTML]{FF0000} \textbf{0.398}} & \textbf{0.398} & 0.400 & 0.400 & 0.411 & 0.406 & 0.407 & 0.398 & 0.559 & 0.511 & 0.418 & 0.414 & 0.418 & 0.404 & 0.556 & 0.523 \\
    & avg & {\color[HTML]{FF0000} \textbf{0.276}} & {\color[HTML]{FF0000} \textbf{0.320}} & 0.278 & \textbf{0.323} & \textbf{0.278} & 0.324 & 0.290 & 0.334 & 0.287 & 0.328 & 0.338 & 0.373 & 0.290 & 0.333 & 0.295 & 0.331 & 0.354 & 0.401 \\
    \midrule
    \multirow{5}{*}{Weather} & 96 & \textbf{0.162} & {\color[HTML]{FF0000} \textbf{0.204}} & 0.162 & \textbf{0.208} & {\color[HTML]{FF0000} \textbf{0.161}} & 0.208 & 0.175 & 0.215 & 0.197 & 0.236 & 0.171 & 0.227 & 0.175 & 0.217 & 0.168 & 0.219 & 0.196 & 0.256 \\
    & 192 & 0.209 & {\color[HTML]{FF0000} \textbf{0.249}} & {\color[HTML]{FF0000} \textbf{0.207}} & \textbf{0.249} & \textbf{0.207} & 0.251 & 0.225 & 0.257 & 0.243 & 0.273 & 0.218 & 0.280 & 0.220 & 0.255 & 0.225 & 0.265 & 0.238 & 0.299 \\
    & 336 & 0.266 & \textbf{0.293} & {\color[HTML]{FF0000} \textbf{0.263}} & {\color[HTML]{FF0000} \textbf{0.290}} & \textbf{0.264} & 0.293 & 0.279 & 0.298 & 0.292 & 0.308 & 0.265 & 0.317 & 0.279 & 0.297 & 0.281 & 0.303 & 0.281 & 0.330 \\
    & 720 & 0.345 & \textbf{0.343} & \textbf{0.338} & {\color[HTML]{FF0000} \textbf{0.340}} & 0.345 & 0.345 & 0.361 & 0.350 & 0.368 & 0.357 & {\color[HTML]{FF0000} \textbf{0.326}} & 0.351 & 0.356 & 0.348 & 0.359 & 0.354 & 0.345 & 0.381 \\
    & avg & 0.246 & \textbf{0.272} & {\color[HTML]{FF0000} \textbf{0.242}} & {\color[HTML]{FF0000} \textbf{0.271}} & \textbf{0.244} & 0.274 & 0.260 & 0.280 & 0.275 & 0.293 & 0.245 & 0.293 & 0.257 & 0.279 & 0.258 & 0.285 & 0.265 & 0.316 \\
    \midrule
    \multirow{5}{*}{ECL} & 96 & \textbf{0.153} & \textbf{0.245} & 0.174 & 0.266 & 0.156 & 0.247 & {\color[HTML]{FF0000} \textbf{0.148}} & {\color[HTML]{FF0000} \textbf{0.240}} & 0.209 & 0.280 & 0.171 & 0.260 & 0.180 & 0.272 & 0.168 & 0.271 & 0.210 & 0.301 \\
    & 192 & \textbf{0.166} & \textbf{0.256} & 0.182 & 0.272 & 0.170 & 0.260 & {\color[HTML]{FF0000} \textbf{0.164}} & {\color[HTML]{FF0000} \textbf{0.256}} & 0.205 & 0.281 & 0.177 & 0.268 & 0.187 & 0.279 & 0.187 & 0.289 & 0.210 & 0.304 \\
    & 336 & \textbf{0.183} & \textbf{0.274} & 0.196 & 0.286 & 0.187 & 0.278 & {\color[HTML]{FF0000} \textbf{0.177}} & {\color[HTML]{FF0000} \textbf{0.270}} & 0.218 & 0.295 & 0.190 & 0.284 & 0.204 & 0.295 & 0.201 & 0.302 & 0.223 & 0.319 \\
    & 720 & {\color[HTML]{FF0000} \textbf{0.221}} & {\color[HTML]{FF0000} \textbf{0.305}} & 0.236 & 0.320 & 0.227 & \textbf{0.312} & \textbf{0.228} & 0.313 & 0.260 & 0.327 & 0.228 & 0.316 & 0.245 & 0.328 & 0.229 & 0.324 & 0.257 & 0.349 \\
    & avg & \textbf{0.181} & \textbf{0.270} & 0.197 & 0.286 & 0.185 & 0.274 & {\color[HTML]{FF0000} \textbf{0.179}} & {\color[HTML]{FF0000} \textbf{0.269}} & 0.223 & 0.296 & 0.191 & 0.282 & 0.204 & 0.294 & 0.196 & 0.297 & 0.225 & 0.318 \\
    \midrule
    \multicolumn{2}{c}{Total AVG} & {\color[HTML]{FF0000} \textbf{0.310}} & {\color[HTML]{FF0000} \textbf{0.345}} & \textbf{0.317} & \textbf{0.352} & 0.325 & 0.355 & 0.330 & 0.359 & 0.337 & 0.359 & 0.351 & 0.381 & 0.328 & 0.361 & 0.338 & 0.368 & 0.379 & 0.403 \\
    \multicolumn{2}{c}{1st Times} & {\color[HTML]{FF0000} \textbf{19}} & {\color[HTML]{FF0000} \textbf{21}} & \textbf{5} & \textbf{4} & 2 & 0 & 4 & 4 & 0 & 2 & 1 & 0 & 0 & 0 & 0 & 0 & 0 & 0 \\
    \bottomrule
    \end{tabular}
  }
  \end{table*}

\subsection{Experiment setting}
Experimental datasets: In order to verify the prediction accuracy of our model on time series generated in real environments, we selected six commonly used real-world datasets: Weather, ETTh1, ETTh2, ETTm1, ETTm2 and Electricity(\cite{zhou_informer_2021,wu_autoformer_2021}) and conducted sufficient experiments on these six datasets to verify the ability of our model in long-term forecasting.

Benchmark models: Based on timeliness, innovation and prediction effect, we selected 8 time series forecasting models which are widely acclaimed in the field of time series forecasting as our baselines, including: (1) TimeKAN(\cite{huang_timekan_2025}) (2) TimeMixer (\cite{wang_timemixer_2024}) (3) iTransformer(\cite{liu_itransformer_2024}) (4) SparseTSF(\cite{lin_sparsetsf_2024}) (5) FreTS(\cite{yi_frequency-domain_2023}) (6) PatchTST(\cite{nie_time_2023}) (7) TimesNet(\cite{wu_timesnet_2023}) (8) DLinear(\cite{zeng_are_2023})

Experimental environment and related indicators: All experiments were implemented based on PyTorch and conducted on a single NVIDIA 3090 24GB GPU. At the same time, in order to ensure fair competition among the models, we set the lookback window, prediction length, and evaluation index to 96, {96, 192, 336, 720}, mean square error (MSE), and mean absolute error (MAE) respectively. What’s more, the benchmark model is tested using the scripts provided in the original code, while the test of the TimeCF model we proposed sets different training rounds and early stopping thresholds according to the size of different data sets to improve test efficiency.

\subsection{Experiment results}
All results in this experiment are obtained after local experiments(except for FreTS whose results are obtained from the original paper) and all results are shown in Table 1. We define that the lower the values of MSE and MAE, the better the model prediction effect. At the same time, the best results are shown in bold red and the second best results are shown in bold black. It is not difficult to see from Table 1 that the TimeCF we proposed has shown good performance on most datasets, except for weather and ECL, where KAN and Transformer model can better handle the autocorrelation dependencies for high-dimensional datasets. Even if it does not achieve the optimal prediction effect in some datasets, the prediction accuracy of TimeCF is not much different from the results achieved by the optimal model. The average values of MSE and MAE increased by 2.2\% and 1.9\% compared with the suboptimal model. And if we look at the number of times the optimal prediction is obtained, TimeCF is far ahead of all models that appeared in the experiments. This proves that TimeCF has accurate and general prediction capabilities on most natural time series.

\subsection{Ablation experiment}
To demonstrate the accuracy of our design and addition of modules, we used three forms of TimeCF models in the ablation implementation to compare with our selected baseline model TimeMixer: (1) TimeCF with the SAMFre module omitted (2) TimeCF with the convolution part omitted and (3) the complete TimeCF. As shown in Table 2, TimeCF without complete modules has a certain improvement over the baseline model in the experiment, but the improvement is not significant. This shows that both the decoupling of label autocorrelation in time series and the enhancement of generalization ability based on SAMFre and the adaptive information aggregation between different scales based on convolution can only enhance the partial information extraction and prediction capabilities of the baseline model TimeMixer to a certain extent. However, the good performance of the complete TimeCF shows that the information of different scales and receptive fields obtained by convolution may contain some information with autocorrelation. And by using SAMFre, the autocorrelation within this part of information can be properly decoupled, which is reflected in the results that it exceeds the baseline model in terms of evaluation indicators. Finally, it is proved that the adaptive information aggregation module based on convolution and the SAMFre module proposed by us are both indispensable parts of the TimeCF model.
\begin{table}[!t]
  \caption{Ablation study of TimeCF. }
  \label{tab:ablation}
  \setlength{\tabcolsep}{3pt}
  \footnotesize
  \begin{tabular}{lcccccc}
  \toprule
  \multirow{2}{*}{Model} & \multicolumn{2}{c}{ETT h1} & \multicolumn{2}{c}{ETT h2} & \multicolumn{2}{c}{ECL} \\
  \cmidrule(lr){2-3} \cmidrule(lr){4-5} \cmidrule(lr){6-7}
   & MSE & MAE & MSE & MAE & MSE & MAE \\
  \midrule
  TimeMixer & 0.469 & 0.449 & 0.392 & 0.411 & 0.185 & 0.274 \\
  TimeCF w/o SAMFre & 0.466 & 0.452 & 0.392 & 0.417 & 0.182 & 0.273 \\
  TimeCF w/o CONV & 0.430 & 0.425 & 0.372 & 0.396 & 0.185 & 0.272 \\
  TimeCF (ours) & {\color[HTML]{FF0000} \textbf{0.417}} & {\color[HTML]{FF0000} \textbf{0.427}} & {\color[HTML]{FF0000} \textbf{0.371}} & {\color[HTML]{FF0000} \textbf{0.396}} & {\color[HTML]{FF0000} \textbf{0.181}} & {\color[HTML]{FF0000} \textbf{0.270}} \\
  \bottomrule
  \end{tabular}
\end{table}

\subsection{Model efficiency}
\begin{table}[!t]
  \caption{Parameter comparison of different time series forecasting models on various datasets.}
  \label{tab:efficiency}
  \footnotesize
  \setlength{\tabcolsep}{3pt}
  \begin{tabular}{lcccccc}
  \toprule
  \multirow{2}{*}{Model} & \multicolumn{6}{c}{Parameters (Number)} \\
  \cmidrule(lr){2-7}
   & ETT h1 & ETT h2 & ETT m1 & ETT m2 & Weather & ECL \\
  \midrule
  TimeMixer & 75.3K & 75.3K & 75.3K & 77.5K & 104K & 104K \\
  iTransformer & 224K & 224K & 224K & 224K & 4.83M & 4.83M \\
  TimesNet & 605K & 1.19M & 4.70M & 1.19M & 1.19M & 150M \\
  SparseTSF & 0.041K & 0.041K & 0.581K & 0.581K & 0.581K & 0.041K \\
  TimeCF (ours) & 125K & 125K & 125K & 275K & 179K & 179K \\
  \bottomrule
  \end{tabular}
\end{table}
In order to verify the efficiency of the TimeCF model we proposed, we set the lookback window and the prediction length to 96 and 96 to test the parameter size of the model. We selected three benchmark models based on the Transformer architecture, the CNN architecture and the MLP model, and a model with the smallest number of parameters as the baseline model for model efficiency. It is not difficult to see from Table 3 that the Transformer and CNN-based models are limited by the model structure, and their parameter volume is maintained at a very high level on all datasets. The model parameters of the MLP-based model TimeMixer and the TimeCF we proposed are basically maintained at a relatively low level on each dataset, and the fluctuation range is not large. Although the number of parameters of SparseTSF is much smaller than that of the TimeCF we proposed and TimeMixer, considering the balance between prediction effect and parameter volume, we believe that the TimeCF we proposed has stable and efficient model operation efficiency while ensuring the accuracy of the prediction results under different datasets. Therefore, it can be considered that the TimeCF we proposed can achieve excellent prediction performance with only a relatively small amount of computing resources.

\section{Conclusion}
In our paper, we proposed a time series prediction model TimeCF based on the TimeMixer decomposition-learning-mixing architecture to achieve high-precision time series forecasting. With the support of PDMC, TimeCF can utilize the information of different receptive fields of sequences of different scales, learn and mix the seasonal and trend sequences separately and finally combine SAMFre to decouple the autocorrelation between labels and reduce the sharpness of the loss function. The performance of our model on real datasets also proves that TimeCF can cope with time series prediction tasks in the real world with good prediction performance.

\bibliographystyle{cas-model2-names}

\bibliography{refernece/TimeCF}

\end{document}